\documentclass[runningheads]{llncs}
\usepackage[T1]{fontenc}
\usepackage[utf8]{inputenc}
\usepackage{graphicx}
\usepackage{booktabs}
\usepackage{multirow}
\usepackage{amsmath,amssymb}
\usepackage{xcolor}
\usepackage{url}
\usepackage{hyperref}
\usepackage[capitalise,noabbrev]{cleveref}
\usepackage{siunitx}

\graphicspath{{figures/}}
\newcommand{\ci}[1]{{\scriptsize$\pm$#1}}
\newcommand{\sat}[1]{S@#1}
\newcommand{\ttoi}{T$\rightarrow$I}
\newcommand{\itot}{I$\rightarrow$T}
\newcommand{\svo}{SVO-Probes}
\newcommand{\qft}{Qwen2.5-VL-FT}
\newcommand{\qemb}{Qwen3-VL-Embed}
\newcommand{\repolink}{\url{https://anonymous.4open.science/r/probe-retrieval-supplementary-4CC4}}
\begin{document}

\title{It Takes Two to Probe, Thousands to Retrieve:\\ Putting Vision--Language Probes\\ to the Retrieval Test}
\titlerunning{It Takes Two to Probe, Thousands to Retrieve}
\author{Ivana Beňová, Michal Gregor, Marián Šimko}
\authorrunning{Beňová et al.}
\institute{Kempelen Institute of Intelligentz Technologies, Bratislava, Slovakia}
\maketitle

\begin{abstract}
Vision--language models (VLMs) are tested for fine-grained understanding with image--text matching probes, a model must prefer the correct image or caption over handcrafted negative. In retrieval, the correct item must beat thousands of alternatives. We measure how these two evaluation methods relate by using two probing benchmarks, \svo{} and SugarCrepe, for retrieval tasks. For 100 image and caption queries we judge the top-10 retrievals from full dataset of six VLMs, with human annotators and an LLM assessor. We than compare how well models rank pool of items on the same queries, while changing only where the negatives comes from: the benchmark, a random draw, or the judged-incorrect retrievals of other models. On \svo{} the accuracy of retrieving a positive over handcrafted negatives is significantly higher than for retrieved ones (the correct item is preferred in 85\% against 77\% of the time). Using pool of ten retrieved negatives leads to same accuracy as when one uses pool of 200 to 430 random negatives. On SugarCrepe the order is reversed: its handcrafted negatives are harder than the retrieved ones. Incomplete relevance labels explain most of the low retrieval scores on \svo{}, where the top-ranked item is correct for 54--90\% of the queries under human labels against 6--21\% under the benchmark's labels. However, we detect no association between passing the probe for a query and retrieving a correct item first. The LLM assessor o3 agrees with individual annotators ($\kappa$ 0.57--0.61) slightly less than they agree with each other (0.64--0.68).

\end{abstract}

\section{Introduction}
\label{sec:intro}

Fine-grained understanding in vision--language models (VLMs) is usually measured with image--text matching (ITM) probes or with similar visual question answering (VQA) tasks. Benchmarks such as \svo{} \cite{hendricks2021probing}, VALSE \cite{parcalabescu2021valse}, Winoground \cite{thrush2022winoground} or SugarCrepe \cite{hsieh2023sugarcrepe} take a correct image--caption pair, add handcrafted negatives that differ in a single semantic component, and report how often the model prefers the correct pair. Probes are cheap to run and easy to interpret, since each negative isolates one phenomenon, and progress on them is read as progress in fine-grained understanding~\cite{bugliarello2023measuring,lewis2024bind}.

\begin{figure}[tb]
\centering
\includegraphics[width=0.8\linewidth]{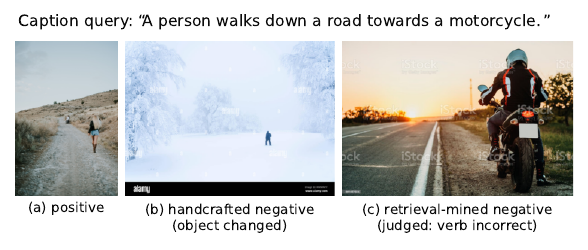}
\caption{A caption query from \svo{}. All six models score the positive (a) above the handcrafted negative (b), in which the object was changed. Image (c) was retrieved in the top 10 and judged incorrect, however, five models score it above the positive.}
\label{fig:example}
\end{figure}

That reading assumes that a probe predicts how the model behaves where fine-grained understanding matters. For dual encoders the main such setting is retrieval. It requires the correct item to be ranked above thousands of alternatives, and the most challenging are the items closest to the correct one in the model's own representation space, not necessarily the ones a benchmark designer chose (\cref{fig:example}). In information retrieval terms, an ITM probe is a ranking evaluation with a candidate set of size two and a fixed distribution of distractors. Whether that distribution resembles the distractors a model meets in retrieval is an open question, and the usual way of answering it, comparing probe accuracy with Recall@1 on a retrieval benchmark, proves little: a two-way choice and a many-way ranking have different chance levels, and relevance labels are incomplete, so a retrieved item that is correct but unlabelled counts as an error.

Prior work has approached the two sides separately. One line focuses on making negatives harder for probing, by filtering them ~\cite{hsieh2023sugarcrepe} or by adding negative images~\cite{miranda2024bivlc,awal2024vismin} or mine natural adversarial examples for new benchmarks~\cite{tong2024mmvp,li2024naturalbench}. Second focuses on retrieval, for example re-judges collections to remove false negatives~\cite{chun2022eccv,parekh2021crisscrossed}. None of them puts the handcrafted negatives of a probe next to the negatives that retrieval actually produces for the same query, and none asks whether passing the probe says anything about retrieval for the query.

We do this for two probing benchmarks, \svo{} and SugarCrepe over COCO, which we treat as retrieval tasks for six VLMs. For every benchmark we build a \emph{judged pool}: for 100 image and caption queries, the top-10 retrievals of every model are labelled as correct or incorrect by human annotators and/or LLM assessor o3. The pools allow a controlled experiment. We fix the query and the decision, ``does the positive score above the negative?'', and change only where the negative comes from: the benchmark's handcrafted negatives, random items of the dataset, or judged-incorrect items retrieved by other models or by itself. We ask (RQ1) how hard handcrafted negatives are compared with retrieval-mined ones, (RQ2) how much of the gap between probe accuracy and retrieval success is due to incomplete labels, (RQ3) whether passing the probe for a query predicts a correct first retrieval for the same query, and (RQ4) whether an LLM assessor can stand in for the human judges that such pools require.

\paragraph{Contributions.}
(i)~A controlled comparison of handcrafted and retrieval-mined negatives on two benchmarks. On \svo{} the handcrafted negatives are significantly easier than judged-incorrect items retrieved by other models; the difference survives restricting the mined negatives to those rejected by all three annotators or by a second assessor, and it grows for shallower pools. On SugarCrepe the handcrafted negatives are the harder ones. How a probe compares with retrieval is thus a property of the individual benchmark and has to be measured.
(ii)~Judged top-10 pools for both benchmarks, with human labels on \svo{}, together with metric definitions, annotation instructions and a pipeline that applies to any probing benchmark.\footnote{\repolink}
(iii)~Evidence on what the probe and the strict retrieval metric each miss: incomplete labels explain most but not all of the low Success@1, no association is detected between passing the probe for a query and retrieving a correct item first, and an LLM assessor agrees with individual annotators slightly less than they agree with each other.

\section{Related Work}
\label{sec:related}

\paragraph{Probes with handcrafted negatives.}
Foiling~\cite{shekhar2017foil} set the recipe: take a correct image--caption pair and build a negative that differs in one controlled aspect. \svo{}~\cite{hendricks2021probing} varies the subject, verb or object; VALSE~\cite{parcalabescu2021valse}, VSR~\cite{liu2023vsr}, VL-CheckList~\cite{zhao2022vlchecklist}, CV-Probes~\cite{benova2024cvprobes} and ViLMA~\cite{kesen2024vilma} cover other phenomena; Winoground~\cite{thrush2022winoground}, CREPE~\cite{ma2023crepe} and ARO~\cite{yuksekgonul2023when} target compositionality. Later work repaired the negatives: SugarCrepe~\cite{hsieh2023sugarcrepe} removed those detectable from text alone, and others added negative \emph{images}~\cite{miranda2024bivlc,awal2024vismin,zhang2024countercurate}. CREPE~\cite{ma2023crepe} and ARO~\cite{yuksekgonul2023when} phrase compositionality as retrieval, but over small candidate sets of generated hard negatives with no human check. All of these evaluate a decision on negatives fixed in advance by their designers, as do the expert-written contrast sets of Gardner et al.~\cite{gardner2020contrast}; we keep those as one condition and add negatives mined by retrievers over the same dataset and verified by an assessor.

\paragraph{Retrieval evaluation with incomplete judgments.}
Image--text retrieval is scored against annotated matches per item although many retrieved items are correct but unlabelled~\cite{hu2019bison}. Crisscrossed Captions~\cite{parekh2021crisscrossed} and ECCV Caption~\cite{chun2022eccv} added human judgments for unlabelled MS-COCO~\cite{lin2014coco} pairs, the latter by pooling models and verifying candidates as in IR test-collection practice~\cite{zobel1998reliable,voorhees2000variations}. We apply depth-10 pooling to a probing benchmark and use the pool as ground truth for a controlled experiment, not only as a metric correction.

\paragraph{Hard negatives and LLM assessors.}
Hard negatives mined from a model's own scores are central to contrastive training~\cite{li2021align,robinson2021hard,yuksekgonul2023when} and to dense retrieval~\cite{karpukhin2020dpr}, where unlabelled positives among the mined items are a known problem~\cite{qu2021rocketqa}; MMVP~\cite{tong2024mmvp} and NaturalBench~\cite{li2024naturalbench} mine natural adversarial examples with models to build harder benchmarks. We use mining for evaluation over the probe's own dataset, with an assessor's label on every mined negative. LLMs are increasingly used to produce relevance labels~\cite{faggioli2023perspectives,thomas2024llms,upadhyay2024umbrela,gu2024survey}, with known failure modes~\cite{alaofi2024fooled} and a live debate on whether they can stand in for human assessors at all~\cite{soboroff2024dont,clarke2024llm}. Multimodal assessors have been validated for ad hoc image search~\cite{yang2024toward} and product search~\cite{hosseini2025retrieve}; we validate two of them for a fine-grained matching task against a majority vote and against individual annotators.

\section{Probing Benchmarks as Retrieval Tasks}
\label{sec:collection}

\paragraph{Benchmarks and definitions.}
\svo{}~\cite{hendricks2021probing} consists of triplets $\langle$caption, positive image, negative image$\rangle$ whose negative image shows a scene that differs from the caption's $\langle subject,verb,object\rangle$ triplet in exactly one role. Our copy has 33,685 triplets over 10,978 captions and 13,285 images, of which 11,455 are the positive image of at least one caption and serve as image queries. SugarCrepe~\cite{hsieh2023sugarcrepe} provides 7,511 items over 1,560 images of COCO val2017~\cite{lin2014coco}, each an image with one of its captions and a fluent negative caption that adds, replaces or swaps an object, an attribute or a relation. The retrieval dataset is all 5,000 val2017 images and their 24,789 distinct captions. In both benchmarks the images are the candidates for text-to-image retrieval (\ttoi{}) and the captions the candidates for image-to-text retrieval (\itot{}). The relevant set $R(q)$ of a query is given by the benchmark: the positive images of a caption, or the captions of an image. \emph{Success@$K$} is the share of queries with an item of $R(q)$ in the top $K$. \emph{Pairwise accuracy} is the share of probe items in which the positive pair scores above the negative one. The \emph{handcrafted negatives} of a caption query on \svo{} are the negative images of its triplets, and those of an image query are the captions of the negative images of its triplets. SugarCrepe's negatives are captions, so its probe is image-to-text only and its handcrafted negatives exist for image queries only.

\paragraph{Models.}
We evaluate six models. Four are used zero-shot as dual encoders: CLIP ViT-L/14~\cite{radford2021learning}, FLAVA~\cite{singh2022flava}, SigLIP2 base~\cite{tschannen2025siglip2} and the first-stage BLIP-2 Q-Former~\cite{li2023blip2}. For BLIP-2 we use the embedding of the first query token without ITM re-ranking, a reduced variant of the published retrieval setting, which takes the maximum over all 32 query tokens and re-ranks with the matching head. The other two models are built on generative Qwen VLMs. \qft{} is our own light adaptation of Qwen2.5-VL-7B-Instruct~\cite{bai2025qwen25vl}: image and caption are each wrapped in a prompt asking the model to express the input ``in one word''~\cite{jiang2023scaling,jiang2024e5v}, the last hidden state is the embedding, and a QLoRA adapter~\cite{dettmers2023qlora} is trained contrastively on 10,000 MS-COCO training pairs for one epoch. \qemb{}~\cite{qwen3vlembedding2026} is a purpose-built embedding model with 2B parameters, used zero-shot. Retrieval and pairwise scores are cosine similarities of the image and caption embeddings, except that on \svo{} CLIP, FLAVA and SigLIP2 are scored with the dot product of their unnormalised embeddings, the rule under which their judged lists were produced (\cref{sec:robust}). Checkpoints and prompts are listed in the repository.

\paragraph{Judged pools.}
The pool queries are 100 caption--image pairs sampled at random: the captions are the caption queries and their positive images the image queries. For each query we retrieved the top 10 candidates of every model: 3,714 unique pairs in each direction on \svo{}, and 3,071 and 3,491 on COCO. A pair is labelled \emph{correct} when the caption could describe the image fully and \emph{incorrect} when its subject, verb or object does not match the image; an incorrect pair carries the error type. Every model's list is labelled separately and blind to the model that produced it. On \svo{}, the \itot{} lists of CLIP, BLIP-2, FLAVA and SigLIP2 were labelled by three annotators (an expert-author, and two volunteers; all have tertiary education, are non-native speakers of English and followed the written instructions that we release) and decided by majority vote; all other \svo{} lists, including the \itot{} lists of the two Qwen models, were labelled by the expert alone. A pair retrieved by several models was labelled once per list. All \svo{} lists were also labelled by the LLM assessor o3. On COCO, the pool was labelled by o3, and the expert labelled a subset to validate it on this benchmark. Agreement between the annotators and with the assessor is reported in \cref{sec:judges}.

\section{Retrieval-Mined Contrast Sets}
\label{sec:method}

Let $q$ be a query with relevant set $R(q)$ and judged pool $P(q)$, and let $P^{-}(q)$ be the pool items judged incorrect; each $c\in P^{-}(q)$ is a negative that a real retriever ranked in its top 10 and an assessor verified to be
wrong. For a system with scoring function $s$, an item of $P^{-}(q)$ is \emph{self-mined} if that model retrieved it in its own top 10 and \emph{other-mined} if only other models did. Self-mined negatives are the model's own mistakes. They are close to or above the positive by selection, so their accuracy is a lower reference point and not a test. Other-mined negatives are items the model never ranked in its top 10, and they are the condition that we compare with the handcrafted negatives.

\paragraph{Matched negatives.}
For each query we evaluate the same decision against four sets of negatives $\mathcal{N}(q)$: (a)~the benchmark's handcrafted negatives, (b)~ten random items of the dataset\footnote{Random negatives are averaged over 20 draws that all models share} $R(q)$, (c)~the other-mined and (d)~the self-mined negatives. We report the share of negatives scored below the best relevant item, $\mathrm{acc}_{\mathcal{N}}(q) = |\mathcal{N}(q)|^{-1}\sum_{c\in\mathcal{N}(q)} \mathbf{1}[\max_{p\in R(q)} s(q,p) > s(q,c)]$, averaged over the queries for which $\mathcal{N}(q)$ is not empty. The query, the relevant items, the scoring function and the decision rule are therefore the same in all four conditions. Condition (b) separates “more candidates” from “harder candidates”; (c) tests whether mined
negatives are hard only for their miner. On the second collection, whose handcrafted negatives are captions, (a) exists for image queries only, and (b) is the reference of the paired tests for caption queries.

\paragraph{$n$-way ranking curves.} For each query, we measure whether the best positive is ranked above all of $k$ negatives, for two types of negatives: $k$ random items of the dataset, $k\in\{1,2,5,\ldots,5000,\text{all}\}$, and $k\le 10$ other-mined negatives. Random negatives are drawn 50 times per query; other-mined negatives are added one by one in a random order, averaged over 50 orders. Both curves are computed on the queries that have at least ten other-mined negatives, so that the set of queries does not change with $k$.

\section{Results}
\label{sec:results}

\paragraph{Probe accuracy and retrieval success.}
\label{sec:gap}
\cref{tab:pairwise} and \cref{tab:full} confirm the picture that motivated this work. Pairwise accuracy on \svo{} is close to 100\%, with object negatives the easiest and verb negatives the hardest, whereas strict Success@1 over the full dataset is very low for caption and also image queries. On the 100 pool queries the top-ranked item is correct under the human labels for 54--90\%, against a strict Success@1 of 6--21\% on the same queries. Incomplete labels thus explain most of the low strict scores, but for 10 to 46\% of the queries the top-ranked item is wrong under human labels too. Whether the probe orders the models as retrieval does cannot be settled with six models. \qemb{} is first on the probe and on full-dataset and human-judged Success@1, and Kendall's $\tau$ between pairwise accuracy and human-judged Success@1 is 0.60 in both directions ($p=0.14$). The clearest disagreement concerns FLAVA and SigLIP2, which are level on the probe, but for image queries the top-ranked caption of SigLIP2 is correct 16 points more often than that of FLAVA ($p=0.007$, paired randomisation test), whereas for caption queries, the difference is 3 points and not significant.

Pairwise accuracy on SugarCrepe is lower than for \svo{}, as the dataset negatives were filtered, with object edits the easiest and attribute and relation edits the hardest. Success@1 is far higher than on \svo{}, because every image has five relevant captions and the dataset has about a third as many images. Every pool-based number for SugarCrepe rests on the o3 labels validated in \cref{sec:judges}, not on human labels. The retrieval numbers of the two Qwen models are in-domain here: \qft{} was adapted on COCO training pairs and the training mix of \qemb{} is likely to include COCO. Under the o3 labels the top-ranked item is correct for 40--93\% of the pool queries, against a strict 22--69\%. Kendall's $\tau$ between pairwise accuracy and Success@1 is 0.87 for caption and 0.47 for image queries ($p=0.02$ and $0.27$, uncorrected), which six models again cannot settle.

\begin{table}[tb]
\centering
\caption{Pairwise accuracy of the six models on \svo{} (33,685 triplets; overall and by the role that the negative changes) and on SugarCrepe (7,511 items; overall and by edit type), with bootstrap 95\% half-widths.}
\label{tab:pairwise}
\footnotesize
\setlength{\tabcolsep}{3pt}
\resizebox{\linewidth}{!}{\begin{tabular}{l cccc cccc}
\toprule
& \multicolumn{4}{c}{SVO-Probes} & \multicolumn{4}{c}{SugarCrepe}\\
\cmidrule(lr){2-5}\cmidrule(lr){6-9}
Model & All & Subj. & Verb & Obj. & All & Obj. & Att. & Rel.\\
\midrule
CLIP & 82.0\ci{0.4} & 84.0\ci{1.0} & 79.9\ci{0.5} & 87.0\ci{0.8} & 79.2\ci{0.9} & 85.0\ci{1.1} & 73.8\ci{1.9} & 71.2\ci{2.5} \\
BLIP-2 & 87.6\ci{0.4} & 90.0\ci{0.8} & 85.3\ci{0.5} & 92.8\ci{0.6} & 77.2\ci{0.9} & 87.3\ci{1.1} & 67.2\ci{2.0} & 64.2\ci{2.5} \\
FLAVA & 86.3\ci{0.4} & 87.4\ci{0.9} & 83.8\ci{0.5} & 93.0\ci{0.6} & 83.8\ci{0.8} & 90.5\ci{0.9} & 77.4\ci{1.7} & 74.7\ci{2.2} \\
SigLIP2 & 86.0\ci{0.4} & 85.1\ci{0.9} & 84.6\ci{0.5} & 91.2\ci{0.7} & 82.7\ci{0.8} & 90.1\ci{0.9} & 77.4\ci{1.8} & 69.9\ci{2.5} \\
Qwen2.5-VL-FT & 87.0\ci{0.4} & 87.1\ci{0.9} & 85.3\ci{0.5} & 91.8\ci{0.6} & 80.9\ci{0.9} & 87.1\ci{1.0} & 74.7\ci{1.9} & 72.5\ci{2.3} \\
Qwen3-VL-Embed & 90.8\ci{0.3} & 93.5\ci{0.6} & 89.1\ci{0.4} & 93.7\ci{0.6} & 87.6\ci{0.7} & 92.1\ci{0.8} & 84.9\ci{1.4} & 79.2\ci{2.1} \\
\bottomrule
\end{tabular}
}
\end{table}

\begin{table}[tb]
\centering
\caption{Retrieval: strict Success@$K$ over the full dataset (\svo{}: 10,978 caption and 11,455 image queries; COCO: 24,789 and 5,000; bootstrap 95\% half-widths), and Success@1 on the 100 pool queries with the top-ranked item judged by the benchmark's labels (Strict) and by the human and o3 labels of the model's own list (97--100 labelled queries per cell). The pool values have 95\% half-widths of 5 to 10 points.}
\label{tab:full}
\footnotesize
\setlength{\tabcolsep}{2.5pt}
\resizebox{\linewidth}{!}{\begin{tabular}{l ccc ccc ccc ccc}
\toprule
& \multicolumn{6}{c}{T$\rightarrow$I} & \multicolumn{6}{c}{I$\rightarrow$T}\\
\cmidrule(lr){2-7}\cmidrule(lr){8-13}
& \multicolumn{3}{c}{Full dataset} & \multicolumn{3}{c}{\sat{1}, pool queries} & \multicolumn{3}{c}{Full dataset} & \multicolumn{3}{c}{\sat{1}, pool queries}\\
\cmidrule(lr){2-4}\cmidrule(lr){5-7}\cmidrule(lr){8-10}\cmidrule(lr){11-13}
Model & \sat{1} & \sat{5} & \sat{10} & Strict & Human & o3 & \sat{1} & \sat{5} & \sat{10} & Strict & Human & o3\\
\midrule
\multicolumn{13}{l}{\emph{SVO-Probes}}\\
CLIP & 8.7\ci{0.5} & 23.8 & 32.8 & 10 & 64 & 65 & 6.5\ci{0.5} & 20.5 & 30.3 & 8 & 54 & 46 \\
BLIP-2 & 9.7\ci{0.5} & 27.1 & 37.9 & 12 & 76 & 70 & 13.2\ci{0.6} & 35.3 & 47.1 & 16 & 75 & 77 \\
FLAVA & 8.6\ci{0.5} & 24.3 & 34.7 & 10 & 72 & 67 & 5.6\ci{0.4} & 17.2 & 24.9 & 6 & 62 & 60 \\
SigLIP2 & 10.4\ci{0.6} & 28.5 & 39.1 & 13 & 75 & 76 & 13.4\ci{0.6} & 35.9 & 48.1 & 21 & 78 & 71 \\
Qwen2.5-VL-FT & 12.2\ci{0.6} & 31.7 & 42.6 & 12 & 70 & 79 & 13.2\ci{0.6} & 35.0 & 46.7 & 13 & 79 & 67 \\
Qwen3-VL-Embed & 14.4\ci{0.7} & 36.9 & 49.0 & 12 & 80 & 84 & 16.8\ci{0.7} & 41.0 & 53.3 & 16 & 90 & 83 \\
\midrule
\multicolumn{13}{l}{\emph{COCO val2017}}\\
CLIP & 36.0\ci{0.6} & 61.0 & 71.0 & 42 & -- & 63 & 57.0\ci{1.4} & 79.8 & 86.9 & 57 & -- & 74 \\
BLIP-2 & 30.3\ci{0.6} & 52.5 & 62.6 & 22 & -- & 40 & 40.4\ci{1.3} & 68.1 & 78.7 & 39 & -- & 67 \\
FLAVA & 41.5\ci{0.6} & 71.3 & 81.7 & 58 & -- & 68 & 48.8\ci{1.4} & 80.1 & 89.9 & 49 & -- & 66 \\
SigLIP2 & 48.5\ci{0.6} & 73.1 & 81.5 & 65 & -- & 76 & 64.2\ci{1.3} & 85.6 & 91.2 & 67 & -- & 78 \\
Qwen2.5-VL-FT & 37.9\ci{0.6} & 63.2 & 73.5 & 44 & -- & 60 & 51.8\ci{1.4} & 76.8 & 84.9 & 51 & -- & 68 \\
Qwen3-VL-Embed & 54.8\ci{0.6} & 79.4 & 86.9 & 61 & -- & 74 & 71.2\ci{1.3} & 89.9 & 94.6 & 69 & -- & 93 \\
\bottomrule
\end{tabular}
}
\end{table}

\paragraph{Matched negatives.}
\label{sec:matchedres}
The upper half of \cref{tab:matched} contains the central results for \svo{}. It presents the share of negatives scored below the best relevant item by origin of negatives. Random items of the dataset are almost never preferred over the positive. The handcrafted negatives, which differ in one role, are 4 to 13 points harder. Judged negatives mined from the top-10 lists of \emph{other} models are harder again, by a further 6 to 11 points, and this difference is significant for every model in both directions. The other-mined condition contains only items the model never ranked in its top 10, so a negative there can outrank the positive only when the positive itself is outside the top 10, which is the case for 40 to 72 of the 100 queries. On those queries the positive beats the handcrafted negatives 79--91\% of the time and the other-mined ones 63--74\%, a gap of 10 to 21 points ($p<0.05$ in all cells). The model's own top-10 errors are scored below the positive only 23--40\% of the time in \ttoi{} and 19--46\% in \itot{}. This condition is circular by construction (\cref{sec:method}).

The lower half of \cref{tab:matched} repeats the matched-negative comparison under the o3 labels for Sugar Crepe. In \itot{}, the direction that has handcrafted negatives, SugarCrepe's negatives are harder than the other-mined negatives for every model and lie within 5 points of all pooled negatives. These values exceed the probe accuracy of \cref{tab:pairwise} because every condition scores a negative against the best relevant item and a COCO image has five relevant captions; on \svo{}, where most queries have one relevant item, the two rules differ by at most 1.3 points. The order is the same under the probe's own rule, with every negative scored against the caption from which the SugarCrepe item was built. Accuracy on the handcrafted negatives is then 69--80\%, which is 8 to 16 points below the other-mined negatives, not significantly different from all pooled negatives, and 14 to 42 points above the self-mined ones. SugarCrepe discards negatives that a text-only model can reject~\cite{hsieh2023sugarcrepe}, which may contribute to their hardness. The two benchmarks also differ in the modality of the negatives, the number of relevant items per query and the assessor, so the contrast with \svo{} is an observation on two benchmarks and not an effect that we can attribute to one cause. One reason for the easy other-mined negatives is visible in Success@10. On COCO the positive is inside the top 10 for 53--89\% and 78--95\% of the pool queries, so few other-mined negatives can outrank it. All pooled negatives still lower every model below the random condition in both directions ($p<0.01$ in every cell), and ten other-mined negatives cost as much accuracy as 130 to 230 random images for caption queries and as 150 to 740 random captions for image queries. Because o3 misses a third of the correct items (\cref{sec:judges}), correct items among its negatives could inflate the hardness of the mined conditions. On the 24 queries that the expert labelled, self-mined accuracy changes by 2.1 points on average in absolute terms (at most 7) with the expert's labels, and self-mined negatives stay 6 to 46 points below other-mined ones for every model; this sample is too small to exclude the effect for the whole pool.

\begin{table}[tb]
\centering
\caption{Matched-negative comparison (100 queries per direction): share of negatives scored below the best relevant item, by origin of the negative (Bench.: the handcrafted negatives; SugarCrepe has them for image queries only). Cells are means over the queries that have such negatives (75--100), with bootstrap 95\% half-widths. Holm-corrected randomisation tests of ``Other'' and ``All'' against ``Bench.''\ (against ``Random'' where there is none) and of ``Self'' against ``Other'': $^{**}$ $p<0.01$, $^{*}$ $p<0.05$.}
\label{tab:matched}
\footnotesize
\setlength{\tabcolsep}{2pt}
\resizebox{\linewidth}{!}{\begin{tabular}{l ccccc ccccc}
\toprule
& \multicolumn{5}{c}{T$\rightarrow$I} & \multicolumn{5}{c}{I$\rightarrow$T}\\
\cmidrule(lr){2-6}\cmidrule(lr){7-11}
Model & Bench. & Random & Other & Self & All & Bench. & Random & Other & Self & All\\
\midrule
\multicolumn{11}{l}{\emph{SVO-Probes (human labels)}}\\
CLIP & 86.7\ci{4.8} & 95.6\ci{2.8} & 77.4\ci{5.9}$^{**}$ & 23.3\ci{8.0}$^{**}$ & 62.1\ci{5.8}$^{**}$ & 87.8\ci{4.6} & 98.7\ci{0.5} & 76.8\ci{5.5}$^{**}$ & 19.4\ci{7.3}$^{**}$ & 61.5\ci{5.6}$^{**}$ \\
BLIP-2 & 90.8\ci{3.6} & 98.7\ci{1.0} & 82.4\ci{5.0}$^{**}$ & 33.4\ci{9.3}$^{**}$ & 71.7\ci{5.4}$^{**}$ & 93.1\ci{3.6} & 99.1\ci{0.8} & 85.2\ci{4.9}$^{**}$ & 41.0\ci{9.9}$^{**}$ & 76.6\ci{5.3}$^{**}$ \\
FLAVA & 90.1\ci{4.2} & 98.8\ci{0.7} & 79.8\ci{5.3}$^{**}$ & 25.8\ci{8.4}$^{**}$ & 66.8\ci{5.5}$^{**}$ & 85.2\ci{5.0} & 98.2\ci{1.1} & 79.0\ci{5.0}$^{*}$ & 23.8\ci{7.7}$^{**}$ & 61.7\ci{5.4}$^{**}$ \\
SigLIP2 & 89.5\ci{4.6} & 98.3\ci{1.2} & 79.5\ci{5.7}$^{**}$ & 37.3\ci{9.6}$^{**}$ & 71.1\ci{6.0}$^{**}$ & 94.6\ci{3.0} & 99.6\ci{0.2} & 87.8\ci{4.9}$^{*}$ & 46.0\ci{9.9}$^{**}$ & 81.0\ci{5.2}$^{**}$ \\
Qwen2.5-VL-FT & 91.1\ci{3.6} & 99.1\ci{0.5} & 83.6\ci{4.8}$^{**}$ & 31.7\ci{9.2}$^{**}$ & 73.3\ci{5.4}$^{**}$ & 93.0\ci{3.2} & 99.4\ci{0.3} & 84.2\ci{4.9}$^{**}$ & 39.9\ci{9.5}$^{**}$ & 74.9\ci{5.4}$^{**}$ \\
Qwen3-VL-Embed & 95.1\ci{2.7} & 99.3\ci{0.4} & 85.4\ci{5.0}$^{**}$ & 40.4\ci{10.3}$^{**}$ & 78.8\ci{5.1}$^{**}$ & 95.7\ci{2.4} & 99.7\ci{0.1} & 88.0\ci{4.6}$^{**}$ & 37.3\ci{10.5}$^{**}$ & 81.4\ci{4.9}$^{**}$ \\
\midrule
\multicolumn{11}{l}{\emph{COCO with SugarCrepe (o3 labels)}}\\
CLIP & -- & 99.7\ci{0.1} & 92.5\ci{3.7}$^{**}$ & 67.0\ci{8.1}$^{**}$ & 84.3\ci{4.5}$^{**}$ & 93.8\ci{3.7} & 100.0\ci{0.0} & 97.4\ci{1.9} & 73.7\ci{8.0}$^{**}$ & 90.3\ci{3.5} \\
BLIP-2 & -- & 99.4\ci{0.2} & 87.1\ci{4.3}$^{**}$ & 40.9\ci{8.3}$^{**}$ & 71.9\ci{5.1}$^{**}$ & 91.2\ci{4.7} & 99.9\ci{0.1} & 98.4\ci{1.0}$^{**}$ & 64.2\ci{8.3}$^{**}$ & 88.8\ci{3.1} \\
FLAVA & -- & 99.9\ci{0.1} & 96.1\ci{3.0}$^{*}$ & 79.3\ci{7.3}$^{**}$ & 91.5\ci{3.6}$^{**}$ & 96.4\ci{2.5} & 100.0\ci{0.0} & 99.7\ci{0.3}$^{**}$ & 74.8\ci{7.3}$^{**}$ & 93.4\ci{2.0} \\
SigLIP2 & -- & 99.9\ci{0.1} & 95.7\ci{2.7}$^{**}$ & 83.8\ci{7.1}$^{**}$ & 92.4\ci{3.7}$^{**}$ & 94.6\ci{3.5} & 100.0\ci{0.0} & 99.2\ci{0.8}$^{**}$ & 84.3\ci{5.8}$^{**}$ & 95.5\ci{1.9} \\
Qwen2.5-VL-FT & -- & 99.7\ci{0.2} & 93.2\ci{3.4}$^{**}$ & 68.9\ci{8.0}$^{**}$ & 86.0\ci{4.5}$^{**}$ & 97.8\ci{1.4} & 100.0\ci{0.0} & 98.0\ci{1.3} & 74.4\ci{7.8}$^{**}$ & 92.8\ci{2.3}$^{**}$ \\
Qwen3-VL-Embed & -- & 99.9\ci{0.1} & 97.1\ci{2.1}$^{*}$ & 84.6\ci{6.4}$^{**}$ & 93.7\ci{2.8}$^{**}$ & 94.7\ci{3.5} & 100.0\ci{0.0} & 99.8\ci{0.3}$^{**}$ & 88.4\ci{6.1}$^{**}$ & 98.1\ci{1.1} \\
\bottomrule
\end{tabular}
}
\end{table}

\paragraph{Robustness of the difference.}
\label{sec:robust}
Four checks address whether the pool or the scoring rule produces the difference between handcrafted and other-mined negatives. The first is label noise, because a mined negative that is in fact correct legitimately scores high. When the other-mined negatives of image queries are restricted to those that all three annotators marked incorrect (51--64\% of them), the handcrafted negatives remain 4 to 8 points easier ($p<0.05$ for four models, $p<0.1$ for BLIP-2 and FLAVA). When the negatives of both directions are restricted to those that o3 also labelled incorrect (81--90\%), the difference is 6 to 9 points and significant in all twelve cells. The second is the two Qwen embedders, whose lists have one annotator: without the items that only they retrieved, the difference is 6 to 12 points ($p<0.05$ in all cells). The third is pool depth. With pools of depth 5 and depth 3 built from the same labels, the difference grows (8--15 points at depth 5 and 11--19 at depth 3, $p<0.01$ in every cell), so it is largest for the highest-ranked retrievals and would probably shrink for pools deeper than 10. The fourth is the scoring rule of CLIP, FLAVA and SigLIP2. Cosine similarity raises their strict Success@1 for image queries from 5.6--13.4\% to 10.0--16.4\%, but on the same judged negatives the difference remains 6 to 11 points ($p<0.05$ in all six cells), and 3 to 9 points ($p<0.05$ in four cells) when the negatives inside their cosine top 10 are removed as well.

\begin{figure}[tb]
\centering
\includegraphics[width=0.76\linewidth]{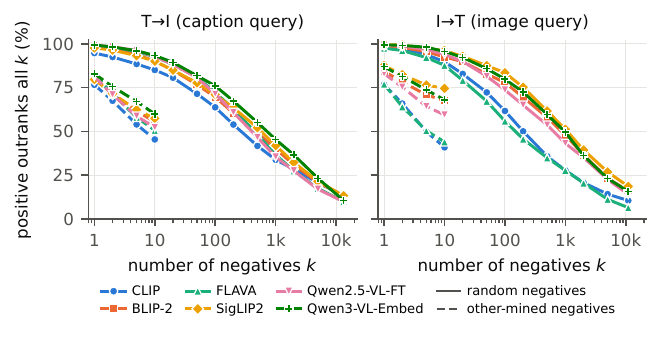}
\caption{Share of queries on which the positive outranks all $k$ negatives on SVO-Probes. Solid: $k$ random items of the dataset, up to the whole dataset. Dashed: $k\le 10$ other-mined negatives. Both curves use the queries with at least ten other-mined negatives (61--76 per model and direction).}
\label{fig:nway}
\end{figure}

\paragraph{Candidate-set size versus semantic hardness.}
\label{sec:nwayres}
\cref{fig:nway} separates the two explanations of low Success@1 on the queries that have at least ten other-mined negatives. Against random negatives the probability that the positive outranks all $k$ of them falls slowly, 85--96\% at $k{=}10$ and 67--87\% at $k{=}50$, and reaches the strict Success@1 level only when $k$ is the whole dataset. Against ten other-mined negatives it is 45--60\% in \ttoi{} and 41--74\% in \itot{}, a level that the random curves reach only at $k$ between 209 and 431. Set size explains the low absolute Success@1, not the gap between the curves.

\paragraph{Probe and retrieval outcome per query.}
\label{sec:joint}
For the same queries we cross-tabulate whether the model scores the positive above all handcrafted negatives of the query, which we call passing the probe and which happens for 65--86\% of the queries, and whether its top-ranked candidate is judged correct. The probe passes while retrieval fails on 14 to 22 caption queries and 8 to 33 image queries, and the opposite occurs on 13 to 21 and 11 to 24 queries. The top-ranked item is correct for 52--91\% of the queries that pass the probe and for 53--86\% of those that fail it, and Fisher's exact test finds no association for any model after correction. With only 14 to 35 queries failing the probe the test would miss moderate effects, so we conclude that no association is detected, not that there is none.

\paragraph{Error types.}
\label{sec:errors}
Between 47\% and 70\% of the items in the judged top-10 lists are correct. Among the incorrect items, verb errors are the largest class for every model in both directions (39--53\% of errors), object errors come second for most models (24--41\%) and subject errors last (11--30\%).

Every pool-based number in this subsection rests on the o3 labels validated in \cref{sec:judges}, not on human labels. The retrieval numbers of the two Qwen models are in-domain here: \qft{} was adapted on COCO training pairs and the training mix of \qemb{} is likely to include COCO.

\paragraph{Matching decisions on the judged pools.}
\label{sec:itmres}
The pools also serve as a test set for the binary form of the probe, in which a model answers whether an image and a caption match and is scored by its accuracy on the benchmark's balanced pairs (ITM accuracy). We re-ran four such decisions on every pool pair of both benchmarks: the matching heads of BLIP-2 and FLAVA, a yes/no prompt to the zero-shot Qwen2.5-VL-7B-Instruct model, and a yes/no construction with \qemb{} (\cref{tab:itmpools}; details in the repository). On the \svo{} pool the two heads reach 54--57\%, against 62--63\% on the benchmark pairs and 48--51\% for a constant answer, and BLIP-2's head answers ``no match'' for 77--87\% of the human-verified correct retrievals. FLAVA's head answers ``no match'' for practically every COCO pair, so its pool accuracy there is the share of incorrect pairs. The \qemb{} construction errs in the other direction and accepts 41--50\% of the incorrect \svo{} pairs, and only the prompted model keeps balanced error rates on both pools. Balanced benchmark pairs thus hide one-sided decisions that the pools expose, which is an observation about these four decisions and not a general result.

\begin{table}[tb]
\centering
\caption{Binary matching decisions: ITM accuracy on the benchmark's own pairs (Bench.) and, on the judged pools, accuracy against the pool labels (Pool), true-positive rate (TPR) and true-negative rate (TNR), for caption / image queries.}
\label{tab:itmpools}
\footnotesize
\setlength{\tabcolsep}{3pt}
\resizebox{\linewidth}{!}{\begin{tabular}{l cccc cccc}
\toprule
& \multicolumn{4}{c}{\svo{} (human labels)} & \multicolumn{4}{c}{COCO with SugarCrepe (o3 labels)}\\
\cmidrule(lr){2-5}\cmidrule(lr){6-9}
Decision & Bench. & Pool & TPR & TNR & Bench. & Pool & TPR & TNR\\
\midrule
BLIP-2 head & 63.4 & 57/54 & 23/13 & 94/94 & 70.4 & 86/77 & 60/75 & 91/79 \\
FLAVA head & 62.4 & 56/54 & 49/46 & 64/62 & 50.0 & 83/59 & 0/0 & 100/100 \\
Qwen2.5-VL prompt & 86.8 & 79/79 & 76/74 & 83/84 & 84.8 & 91/85 & 81/84 & 93/85 \\
\qemb{} yes/no & 84.2 & 75/71 & 90/92 & 59/50 & 76.7 & 79/71 & 93/94 & 77/55 \\
\midrule
Always ``no match'' & 50.0 & 48/51 & 0/0 & 100/100 & 50.0 & 83/59 & 0/0 & 100/100 \\
\bottomrule
\end{tabular}
}
\end{table}

\section{LLM Assessors as a Substitute for Human Judgments}
\label{sec:judges}

Human-judged pools are expensive, so we ask how well an LLM assessor approximates them. The assessors are validated on the \svo{} lists of the four dual encoders: the \itot{} lists, labelled by three annotators (4,000 list entries), and the \ttoi{} lists, labelled by the expert (4,000 entries). Between the three annotators, pairwise Cohen's $\kappa$ on the binary label is 0.64--0.68 and Fleiss' $\kappa$ is 0.64.

\paragraph{LLM assessors on SVO-Probes.}
We prompted GPT-4o and the reasoning model o3~\cite{openai2025o3} (version \texttt{o3-2025-04-16}, medium reasoning effort) with the query and its ten candidates and asked for a label per candidate (\emph{correct} or the error type). \cref{tab:judge} reports agreement on the entries for which the assessor returned a parseable label; 1--4\% of the entries are without one for o3 and 7--14\% for GPT-4o. On the \itot{} half o3 agrees with the human majority on 82.6\% of the entries (F1 0.84 on the \emph{correct} class, Cohen's $\kappa$ 0.65); on the \ttoi{} half, against the expert, on 83.0\% (F1 0.85, $\kappa$ 0.66). GPT-4o is markedly worse, mostly because it misses correct candidates. Agreement with a majority vote overstates agreement with a person, so the comparison on equal terms is with single annotators: o3 reaches $\kappa$ 0.57--0.61 against each of the three, slightly below the 0.64--0.68 that they reach with each other. On the lists of the two Qwen models, labelled by the expert alone, o3 reaches $\kappa$ 0.60--0.68. Error types are harder ($\kappa$ 0.55--0.62 on the four-way label), with most confusion between verb and object errors. The assessor o3 is thus close to an additional annotator without being interchangeable with one~\cite{thomas2024llms,voorhees2000variations}, and we use the human labels as ground truth for the \svo{} experiments.

\begin{table}[tb]
\centering
\caption{Agreement of LLM assessors with the human labels: accuracy, precision and recall of the \emph{correct} class, its F1, Cohen's $\kappa$ on the binary and on the four-way label (correct, subject, verb, object error); $n$ = entries with a parseable label (of 4,000 per half of the \svo{} pool and of 709 on COCO); 95\% intervals from a bootstrap over queries.}
\label{tab:judge}
\footnotesize
\setlength{\tabcolsep}{3pt}
\resizebox{\linewidth}{!}{\begin{tabular}{ll r c c c c c}
\toprule
Assessor & Human labels & $n$ & Acc.\ (\%) & P / R (correct) & F1 (correct) [95\% CI] & $\kappa$ [95\% CI] & $\kappa$ (4-way)\\
\midrule
o3 & \itot{} (majority of 3) & 3960 & 82.6 & 0.88 / 0.80 & 0.84 [0.81, 0.86] & 0.65 [0.60, 0.70] & 0.55 \\
o3 & \ttoi{} (expert) & 3836 & 83.0 & 0.90 / 0.80 & 0.85 [0.82, 0.87] & 0.66 [0.61, 0.71] & 0.62 \\
GPT-4o & \itot{} (majority of 3) & 3729 & 72.8 & 0.82 / 0.66 & 0.73 [0.70, 0.75] & 0.46 [0.42, 0.50] & 0.38 \\
GPT-4o & \ttoi{} (expert) & 3456 & 75.9 & 0.82 / 0.76 & 0.79 [0.76, 0.81] & 0.51 [0.47, 0.55] & 0.46 \\
o3 & COCO sample (expert) & 705 & 88.8 & 0.83 / 0.65 & 0.73 [0.60, 0.81] & 0.66 [0.53, 0.76] & 0.42 \\
\bottomrule
\end{tabular}
}
\end{table}

\paragraph{LLM assessor on COCO.}
The COCO pool of \cref{sec:coco} was labelled by o3 under the same prompt, one call per query over all pooled candidates: 200 calls, 6,554 of 6,562 pairs labelled, about 6 USD. The expert labelled 709 pairs of 12 queries per direction, drawn at random and shown blind to the retrieving models (last row of \cref{tab:judge}). o3 agrees with the expert on 88.8\% of them, with $\kappa$ 0.66 and a wide interval because the sample has 24 queries ($\kappa$ 0.55 for caption and 0.65 for image queries). It is more conservative here than on \svo{}: its precision on the \emph{correct} class is 0.83, but it misses a third of the pairs that the expert accepts (recall 0.65). The COCO results of \cref{sec:coco} therefore rest on a stricter notion of correctness than the \svo{} results.

\section{Limitations and Conclusion}
\label{sec:conclusion}

\paragraph{Limitations.}
Human judgments cover only \svo{}, and the \ttoi{} lists and all lists of the two Qwen models have a single annotator, who is one of the authors; the controls of \cref{sec:robust} address label noise for the central comparison only. Every pool-based number on COCO is an o3 label, validated against 709 expert labels from 24 queries. The two benchmarks differ in more than the construction of their negatives, so the contrast between them is not causal evidence. BLIP-2 is evaluated in a reduced variant, and the two Qwen models are in-domain on COCO. The pools cover 100 queries per direction at depth 10, and the two directions use linked query samples and are not independent replications; they are a diagnostic sample and not reusable for models outside the pool: if the items that only one model contributed (29--62\% of its top 10) were unjudged, its judged Success@1 would drop by 6 to 23 points. Six models are too few for conclusions about model rankings.

\paragraph{Conclusion.} We studied how image--text matching probes relate to retrieval by treating two probing benchmarks as retrieval tasks and comparing negatives of different origin under a fixed query, scoring function and decision rule. On \svo{}, judged-incorrect items retrieved by other models were significantly harder than the handcrafted negatives for all six models. This held under human labels, under labels on which all annotators or two assessors agree, and at every pool depth we examined, and ten such items reduced accuracy as much as two to four hundred random ones. On SugarCrepe the handcrafted negatives were the harder ones. A high probe accuracy therefore does not by itself tell how a model fares against the candidates that retrieval brings up; the relation has to be measured for each benchmark, and judged top-10 pools are sufficient to measure it. The strict retrieval metric understates the models, since most top-ranked items that it counts as errors are correct under human labels, and we detected no association between passing the probe for a query and retrieving a correct item first. For scaling such pools, the LLM assessor o3 came close to the agreement between human annotators without reaching it and was more conservative on COCO than on \svo{}, so an assessor should be validated on each benchmark and against individual annotators.

\bibliographystyle{splncs04}
\bibliography{refs}

@inproceedings{hendricks2021probing,
  title={Probing image-language transformers for verb understanding},
  author={Hendricks, Lisa Anne and Nematzadeh, Aida},
  booktitle={Findings of the Association for Computational Linguistics: ACL-IJCNLP 2021},
  pages={3635--3644},
  year={2021}
}

@inproceedings{thrush2022winoground,
  title={Winoground: Probing vision and language models for visio-linguistic compositionality},
  author={Thrush, Tristan and Jiang, Ryan and Bartolo, Max and Singh, Amanpreet and Williams, Adina and Kiela, Douwe and Ross, Candace},
  booktitle={Proceedings of the IEEE/CVF Conference on Computer Vision and Pattern Recognition},
  pages={5238--5248},
  year={2022}
}

@inproceedings{hsieh2023sugarcrepe,
  title={SugarCrepe: Fixing hackable benchmarks for vision-language compositionality},
  author={Hsieh, Cheng-Yu and Zhang, Jieyu and Ma, Zixian and Kembhavi, Aniruddha and Krishna, Ranjay},
  booktitle={Advances in Neural Information Processing Systems},
  volume={36},
  pages={31096--31116},
  year={2023}
}

@inproceedings{parcalabescu2021valse,
  title={{VALSE}: A task-independent benchmark for vision and language models centered on linguistic phenomena},
  author={Parcalabescu, Letitia and Cafagna, Michele and Muradjan, Lilitta and Frank, Anette and Calixto, Iacer and Gatt, Albert},
  booktitle={Proceedings of the 60th Annual Meeting of the Association for Computational Linguistics (Volume 1: Long Papers)},
  pages={8253--8280},
  year={2022}
}

@inproceedings{ma2023crepe,
  title={{CREPE}: Can vision-language foundation models reason compositionally?},
  author={Ma, Zixian and Hong, Jerry and Gul, Mustafa Omer and Gandhi, Mona and Gao, Irena and Krishna, Ranjay},
  booktitle={Proceedings of the IEEE/CVF Conference on Computer Vision and Pattern Recognition},
  pages={10910--10921},
  year={2023}
}

@inproceedings{kesen2024vilma,
  title={{ViLMA}: A zero-shot benchmark for linguistic and temporal grounding in video-language models},
  author={Kesen, Ilker and Pedrotti, Andrea and Dogan, Mustafa and Cafagna, Michele and Acikgoz, Emre Can and Parcalabescu, Letitia and Calixto, Iacer and Frank, Anette and Gatt, Albert and Erdem, Aykut and Erdem, Erkut},
  booktitle={Proceedings of the 12th International Conference on Learning Representations (ICLR)},
  year={2024}
}

@article{benova2024cvprobes,
  title={{CV-Probes}: Studying the interplay of lexical and world knowledge in visually grounded verb understanding},
  author={Be{\v{n}}ov{\'a}, Ivana and Gregor, Michal and Gatt, Albert},
  journal={arXiv preprint arXiv:2409.01389},
  year={2024}
}

@inproceedings{shekhar2017foil,
  title={{FOIL} it! {F}ind one mismatch between image and language caption},
  author={Shekhar, Ravi and Pezzelle, Sandro and Klimovich, Yauhen and Herbelot, Aur{\'e}lie and Nabi, Moin and Sangineto, Enver and Bernardi, Raffaella},
  booktitle={Proceedings of the 55th Annual Meeting of the Association for Computational Linguistics (Volume 1: Long Papers)},
  pages={255--265},
  year={2017}
}

@article{liu2023vsr,
  title={Visual spatial reasoning},
  author={Liu, Fangyu and Emerson, Guy and Collier, Nigel},
  journal={Transactions of the Association for Computational Linguistics},
  volume={11},
  pages={635--651},
  year={2023}
}

@article{zhao2022vlchecklist,
  title={{VL-CheckList}: Evaluating pre-trained vision-language models with objects, attributes and relations},
  author={Zhao, Tiancheng and Zhang, Tianqi and Zhu, Mingwei and Shen, Haozhan and Lee, Kyusong and Lu, Xiaopeng and Yin, Jianwei},
  journal={arXiv preprint arXiv:2207.00221},
  year={2022}
}

@inproceedings{bugliarello2023measuring,
  title={Measuring progress in fine-grained vision-and-language understanding},
  author={Bugliarello, Emanuele and Sartran, Laurent and Agrawal, Aishwarya and Hendricks, Lisa Anne and Nematzadeh, Aida},
  booktitle={Proceedings of the 61st Annual Meeting of the Association for Computational Linguistics (Volume 1: Long Papers)},
  pages={1559--1582},
  year={2023}
}

@inproceedings{yuksekgonul2023when,
  title={When and why vision-language models behave like bags-of-words, and what to do about it?},
  author={Yuksekgonul, Mert and Bianchi, Federico and Kalluri, Pratyusha and Jurafsky, Dan and Zou, James},
  booktitle={The Eleventh International Conference on Learning Representations},
  year={2023}
}

@inproceedings{miranda2024bivlc,
  title={{BiVLC}: Extending vision-language compositionality evaluation with text-to-image retrieval},
  author={Miranda, Imanol and Salaberria, Ander and Agirre, Eneko and Azkune, Gorka},
  booktitle={Advances in Neural Information Processing Systems},
  volume={37},
  pages={101880--101904},
  year={2024}
}

@inproceedings{radford2021learning,
  title={Learning transferable visual models from natural language supervision},
  author={Radford, Alec and Kim, Jong Wook and Hallacy, Chris and Ramesh, Aditya and Goh, Gabriel and Agarwal, Sandhini and Sastry, Girish and Askell, Amanda and Mishkin, Pamela and Clark, Jack and others},
  booktitle={International Conference on Machine Learning},
  pages={8748--8763},
  year={2021},
  organization={PMLR}
}

@inproceedings{singh2022flava,
  title={{FLAVA}: A foundational language and vision alignment model},
  author={Singh, Amanpreet and Hu, Ronghang and Goswami, Vedanuj and Couairon, Guillaume and Galuba, Wojciech and Rohrbach, Marcus and Kiela, Douwe},
  booktitle={Proceedings of the IEEE/CVF Conference on Computer Vision and Pattern Recognition},
  pages={15638--15650},
  year={2022}
}

@article{tschannen2025siglip2,
  title={{SigLIP 2}: Multilingual vision-language encoders with improved semantic understanding, localization, and dense features},
  author={Tschannen, Michael and Gritsenko, Alexey and Wang, Xiao and Naeem, Muhammad Ferjad and Alabdulmohsin, Ibrahim and Parthasarathy, Nikhil and Evans, Talfan and Beyer, Lucas and Xia, Ye and Mustafa, Basil and others},
  journal={arXiv preprint arXiv:2502.14786},
  year={2025}
}

@inproceedings{li2023blip2,
  title={{BLIP-2}: Bootstrapping language-image pre-training with frozen image encoders and large language models},
  author={Li, Junnan and Li, Dongxu and Savarese, Silvio and Hoi, Steven},
  booktitle={International Conference on Machine Learning},
  pages={19730--19742},
  year={2023},
  organization={PMLR}
}

@article{bai2025qwen25vl,
  title={{Qwen2.5-VL} technical report},
  author={Bai, Shuai and Chen, Keqin and Liu, Xuejing and Wang, Jialin and Ge, Wenbin and Song, Sibo and Dang, Kai and Wang, Peng and Wang, Shijie and Tang, Jun and others},
  journal={arXiv preprint arXiv:2502.13923},
  year={2025}
}

@article{jiang2023scaling,
  title={Scaling sentence embeddings with large language models},
  author={Jiang, Ting and Huang, Shaohan and Luan, Zhongzhi and Wang, Deqing and Zhuang, Fuzhen},
  journal={arXiv preprint arXiv:2307.16645},
  year={2023}
}

@article{jiang2024e5v,
  title={{E5-V}: Universal embeddings with multimodal large language models},
  author={Jiang, Ting and Song, Minzheng and Zhang, Zihan and Huang, Haizhen and Deng, Weiwei and Sun, Feng and Zhang, Qi and Wang, Deqing and Zhuang, Fuzhen},
  journal={arXiv preprint arXiv:2407.12580},
  year={2024}
}

@inproceedings{lin2014coco,
  title={Microsoft {COCO}: Common objects in context},
  author={Lin, Tsung-Yi and Maire, Michael and Belongie, Serge and Hays, James and Perona, Pietro and Ramanan, Deva and Doll{\'a}r, Piotr and Zitnick, C Lawrence},
  booktitle={European Conference on Computer Vision},
  pages={740--755},
  year={2014},
  organization={Springer}
}

@inproceedings{chun2022eccv,
  title={{ECCV} caption: Correcting false negatives by collecting machine-and-human-verified image-caption associations for {MS-COCO}},
  author={Chun, Sanghyuk and Kim, Wonjae and Park, Song and Chang, Minsuk and Oh, Seong Joon},
  booktitle={European Conference on Computer Vision},
  pages={1--19},
  year={2022},
  organization={Springer}
}

@inproceedings{hu2019bison,
  title={Evaluating text-to-image matching using binary image selection ({BISON})},
  author={Hu, Hexiang and Misra, Ishan and Van Der Maaten, Laurens},
  booktitle={Proceedings of the IEEE/CVF International Conference on Computer Vision Workshops},
  year={2019}
}

@inproceedings{zobel1998reliable,
  title={How reliable are the results of large-scale information retrieval experiments?},
  author={Zobel, Justin},
  booktitle={Proceedings of the 21st Annual International ACM SIGIR Conference on Research and Development in Information Retrieval},
  pages={307--314},
  year={1998}
}

@inproceedings{faggioli2023perspectives,
  title={Perspectives on large language models for relevance judgment},
  author={Faggioli, Guglielmo and Dietz, Laura and Clarke, Charles LA and Demartini, Gianluca and Hagen, Matthias and Hauff, Claudia and Kando, Noriko and Kanoulas, Evangelos and Potthast, Martin and Stein, Benno and Wachsmuth, Henning},
  booktitle={Proceedings of the 2023 ACM SIGIR International Conference on Theory of Information Retrieval},
  pages={39--50},
  year={2023}
}

@inproceedings{thomas2024llms,
  title={Large language models can accurately predict searcher preferences},
  author={Thomas, Paul and Spielman, Seth and Craswell, Nick and Mitra, Bhaskar},
  booktitle={Proceedings of the 47th International ACM SIGIR Conference on Research and Development in Information Retrieval},
  pages={1930--1940},
  year={2024}
}

@article{upadhyay2024umbrela,
  title={{UMBRELA}: {UMbrela} is the ({O}pen-source reproduction of the) {B}ing {RELevance} {A}ssessor},
  author={Upadhyay, Shivani and Pradeep, Ronak and Thakur, Nandan and Craswell, Nick and Lin, Jimmy},
  journal={arXiv preprint arXiv:2406.06519},
  year={2024}
}

@inproceedings{alaofi2024fooled,
  title={{LLMs} can be fooled into labelling a document relevant},
  author={Alaofi, Marwah and Thomas, Paul and Scholer, Falk and Sanderson, Mark},
  booktitle={Proceedings of the 2024 Annual International ACM SIGIR Conference on Research and Development in Information Retrieval in the Asia Pacific Region},
  pages={32--41},
  year={2024}
}

@article{gu2024survey,
  title={A survey on {LLM}-as-a-judge},
  author={Gu, Jiawei and Jiang, Xuhui and Shi, Zhichao and Tan, Hexiang and Zhai, Xuehao and Xu, Chengjin and Li, Wei and Shen, Yinghan and Ma, Shengjie and Liu, Honghao and others},
  journal={arXiv preprint arXiv:2411.15594},
  year={2024}
}

@article{openai2025o3,
  title={{OpenAI} o3 and o4-mini system card},
  author={{OpenAI}},
  journal={\url{https://openai.com/index/o3-o4-mini-system-card/}},
  year={2025}
}

@inproceedings{li2021align,
  title={Align before fuse: Vision and language representation learning with momentum distillation},
  author={Li, Junnan and Selvaraju, Ramprasaath and Gotmare, Akhilesh and Joty, Shafiq and Xiong, Caiming and Hoi, Steven Chu Hong},
  booktitle={Advances in Neural Information Processing Systems},
  volume={34},
  pages={9694--9705},
  year={2021}
}

@inproceedings{dettmers2023qlora,
  title={{QLoRA}: Efficient finetuning of quantized {LLMs}},
  author={Dettmers, Tim and Pagnoni, Artidoro and Holtzman, Ari and Zettlemoyer, Luke},
  booktitle={Advances in Neural Information Processing Systems},
  volume={36},
  pages={10088--10115},
  year={2023}
}

@article{voorhees2000variations,
  title={Variations in relevance judgments and the measurement of retrieval effectiveness},
  author={Voorhees, Ellen M},
  journal={Information Processing \& Management},
  volume={36},
  number={5},
  pages={697--716},
  year={2000}
}

@inproceedings{robinson2021hard,
  title={Contrastive learning with hard negative samples},
  author={Robinson, Joshua and Chuang, Ching-Yao and Sra, Suvrit and Jegelka, Stefanie},
  booktitle={International Conference on Learning Representations},
  year={2021}
}

@article{qwen3vlembedding2026,
  author  = {Mingxin Li and Yanzhao Zhang and Dingkun Long and Keqin Chen and Sibo Song and Shuai Bai and Zhibo Yang and Pengjun Xie and An Yang and Dayiheng Liu and Jingren Zhou and Junyang Lin},
  title   = {{Qwen3-VL-Embedding} and {Qwen3-VL-Reranker}: A Unified Framework for State-of-the-Art Multimodal Retrieval and Ranking},
  journal = {arXiv preprint arXiv:2601.04720},
  year    = {2026}
}

@article{soboroff2024dont,
  author  = {Ian Soboroff},
  title   = {Don't Use {LLMs} to Make Relevance Judgments},
  journal = {arXiv preprint arXiv:2409.15133},
  year    = {2024}
}

@article{clarke2024llm,
  author  = {Charles L. A. Clarke and Laura Dietz},
  title   = {{LLM}-based Relevance Assessment Still Can't Replace Human Relevance Assessment},
  journal = {arXiv preprint arXiv:2412.17156},
  year    = {2024}
}

@inproceedings{awal2024vismin,
  author    = {Rabiul Awal and Saba Ahmadi and Le Zhang and Aishwarya Agrawal},
  title     = {{VisMin}: Visual Minimal-Change Understanding},
  booktitle = {Advances in Neural Information Processing Systems (NeurIPS)},
  year      = {2024},
  note      = {arXiv:2407.16772}
}

@inproceedings{zhang2024countercurate,
  author    = {Jianrui Zhang and Mu Cai and Tengyang Xie and Yong Jae Lee},
  title     = {{CounterCurate}: Enhancing Physical and Semantic Visio-Linguistic Compositional Reasoning via Counterfactual Examples},
  booktitle = {Findings of the Association for Computational Linguistics: ACL 2024},
  year      = {2024},
  note      = {arXiv:2402.13254}
}

@article{yang2024toward,
  author  = {Jheng-Hong Yang and Jimmy Lin},
  title   = {Toward Automatic Relevance Judgment Using Vision-Language Models for Image-Text Retrieval Evaluation},
  journal = {arXiv preprint arXiv:2408.01363},
  year    = {2024}
}

@inproceedings{hosseini2025retrieve,
  author    = {Kasra Hosseini and Thomas Kober and Josip Krapac and Roland Vollgraf and Weiwei Cheng and Ana Peleteiro Ramallo},
  title     = {Retrieve, Annotate, Evaluate, Repeat: Leveraging Multimodal {LLMs} for Large-Scale Product Retrieval Evaluation},
  booktitle = {Advances in Information Retrieval: 47th European Conference on Information Retrieval (ECIR)},
  series    = {LNCS},
  volume    = {15572},
  publisher = {Springer},
  year      = {2025}
}

@inproceedings{tong2024mmvp,
  title={Eyes wide shut? {E}xploring the visual shortcomings of multimodal {LLMs}},
  author={Tong, Shengbang and Liu, Zhuang and Zhai, Yuexiang and Ma, Yi and LeCun, Yann and Xie, Saining},
  booktitle={Proceedings of the IEEE/CVF Conference on Computer Vision and Pattern Recognition},
  year={2024}
}

@inproceedings{li2024naturalbench,
  title={{NaturalBench}: Evaluating vision-language models on natural adversarial samples},
  author={Li, Baiqi and Lin, Zhiqiu and Peng, Wenxuan and Nyandwi, Jean de Dieu and Jiang, Daniel and Ma, Zixian and Khanuja, Simran and Krishna, Ranjay and Neubig, Graham and Ramanan, Deva},
  booktitle={Advances in Neural Information Processing Systems},
  volume={37},
  year={2024}
}

@inproceedings{karpukhin2020dpr,
  title={Dense passage retrieval for open-domain question answering},
  author={Karpukhin, Vladimir and O{\u{g}}uz, Barlas and Min, Sewon and Lewis, Patrick and Wu, Ledell and Edunov, Sergey and Chen, Danqi and Yih, Wen-tau},
  booktitle={Proceedings of the 2020 Conference on Empirical Methods in Natural Language Processing},
  pages={6769--6781},
  year={2020}
}

@inproceedings{gardner2020contrast,
  title={Evaluating models' local decision boundaries via contrast sets},
  author={Gardner, Matt and Artzi, Yoav and Basmov, Victoria and Berant, Jonathan and Bogin, Ben and Chen, Sihao and Dasigi, Pradeep and Dua, Dheeru and Elazar, Yanai and Gottumukkala, Ananth and Gupta, Nitish and Hajishirzi, Hannaneh and Ilharco, Gabriel and Khashabi, Daniel and Lin, Kevin and Liu, Jiangming and Liu, Nelson F. and Mulcaire, Phoebe and Ning, Qiang and Singh, Sameer and Smith, Noah A. and Subramanian, Sanjay and Tsarfaty, Reut and Wallace, Eric and Zhang, Ally and Zhou, Ben},
  booktitle={Findings of the Association for Computational Linguistics: EMNLP 2020},
  pages={1307--1323},
  year={2020}
}

@inproceedings{parekh2021crisscrossed,
  title={Crisscrossed captions: Extended intramodal and intermodal semantic similarity judgments for {MS-COCO}},
  author={Parekh, Zarana and Baldridge, Jason and Cer, Daniel and Waters, Austin and Yang, Yinfei},
  booktitle={Proceedings of the 16th Conference of the European Chapter of the Association for Computational Linguistics: Main Volume},
  pages={2855--2870},
  year={2021}
}

@inproceedings{qu2021rocketqa,
  title={{RocketQA}: An optimized training approach to dense passage retrieval for open-domain question answering},
  author={Qu, Yingqi and Ding, Yuchen and Liu, Jing and Liu, Kai and Ren, Ruiyang and Zhao, Wayne Xin and Dong, Daxiang and Wu, Hua and Wang, Haifeng},
  booktitle={Proceedings of the 2021 Conference of the North American Chapter of the Association for Computational Linguistics: Human Language Technologies},
  pages={5835--5847},
  year={2021}
}

@inproceedings{lewis2024bind,
  title={Does {CLIP} bind concepts? {P}robing compositionality in large image models},
  author={Lewis, Martha and Nayak, Nihal and Yu, Peilin and Merullo, Jack and Yu, Qinan and Bach, Stephen and Pavlick, Ellie},
  booktitle={Findings of the Association for Computational Linguistics: EACL 2024},
  pages={1487--1500},
  year={2024}
}

\end{document}